\documentclass{ecai}
\usepackage{graphicx}
\usepackage{latexsym}

\usepackage{times}
\usepackage{soul}
\usepackage{url}
\usepackage[hidelinks]{hyperref}
\usepackage[utf8]{inputenc}
\usepackage[small]{caption}
\usepackage{graphicx}
\usepackage{amsmath}
\usepackage{amsthm}
\usepackage{booktabs}
\usepackage{algorithm}
\usepackage{algorithmic}
\usepackage{multirow}
\usepackage{xcolor}
\usepackage[misc]{ifsym}
\usepackage[symbol]{footmisc}
\usepackage{amsfonts,amssymb}

\begin{document}

\begin{frontmatter}

\title{S2M: Converting Single-Turn to Multi-Turn Datasets for \\
Conversational Question Answering}

\author{Baokui Li$^{a;\dag}$, Sen Zhang$^{a;\dag}$, Wangshu Zhang$^{b}$, Yicheng Chen$^{b}$, Changlin Yang$^{b}$, Sen Hu$^{b}$, Teng Xu$^{b}$, Siye liu$^{b}$, and Jiwei Li$^{c;*}$\\
$^a$School of Software Technology, Zhejiang University \\
$^b$Ant Group \\
$^c$College of Computer Science and Technology, Zhejiang University\vspace{-3ex}}



\renewcommand{\thefootnote}{\arabic{footnote}}

\begin{abstract}
Supplying data augmentation to conversational question answering (CQA) can effectively improve model performance. However, there is less improvement from single-turn datasets in CQA due to the distribution gap between single-turn and multi-turn datasets. 
On the other hand, while numerous single-turn datasets are available, we have not utilized them effectively. To solve this problem, we propose a novel method to convert single-turn datasets to multi-turn datasets. The proposed method consists of three parts, namely, a QA pair Generator, a QA pair Reassembler, and a question Rewriter. Given a sample consisting of context and single-turn QA pairs, the Generator obtains candidate QA pairs and a knowledge graph based on the context. The Reassembler utilizes the knowledge graph to get sequential QA pairs, and the Rewriter rewrites questions from a conversational perspective to obtain a multi-turn dataset S2M. Our experiments show that our method can synthesize effective training resources for CQA. Notably, S2M ranks 1st place on the QuAC leaderboard\footnotemark[1] at the time of submission (Aug 24th, 2022).
\end{abstract}
\end{frontmatter}

\renewcommand{\thefootnote}{\fnsymbol{footnote}}
\footnotetext[2]{Equal contribution.}
\footnotetext[1]{Corresponding Author. Email: jiwei\_li@zju.edu.cn.}

\renewcommand{\thefootnote}{\arabic{footnote}}
\footnotetext[1]{https://quac.ai/}

\vspace{-0.2cm}
\section{Introduction}
    The task of conversational question answering, which requires machines to answer questions through reading and understanding a given context and history Question-Answer pairs, has been a rapidly growing area in natural language understanding~\cite{Choi_QuAC,rajpurkar2018,Rajpurkar_SQuAD,reddy2019coqa}. With the development of pre-trained language models~\cite{Clark2020ELECTRA,he2023debertav,liu2019roberta}, the upper limit of CQA is constantly broken. However, they are still limited by the scale of real-world datasets. More annotated datasets are needed to promote conversational question answering development.

    


    To alleviate the limitation of data scale, mainstream research has explored two methods. 
For example, there are a large number of single-turn datasets~\cite{NQ_2019,PAQ_2021,Rajpurkar_SQuAD} in reading comprehension. While numerous single-turn datasets are available, there are some aspects that have not been fully exploited. Recent studies have shown that the distribution shift between datasets severely constrains only using single-turn datasets. The second method is to automatically generate label datasets~\cite{simseek2022,RGX2022}, whic can generate datasets according to the distribution of target domains. However, generating CQA datasets is challenging and requires interaction between question-answer pairs.

\begin{figure}[htb]
    \centering
    \includegraphics[width=0.5\textwidth]{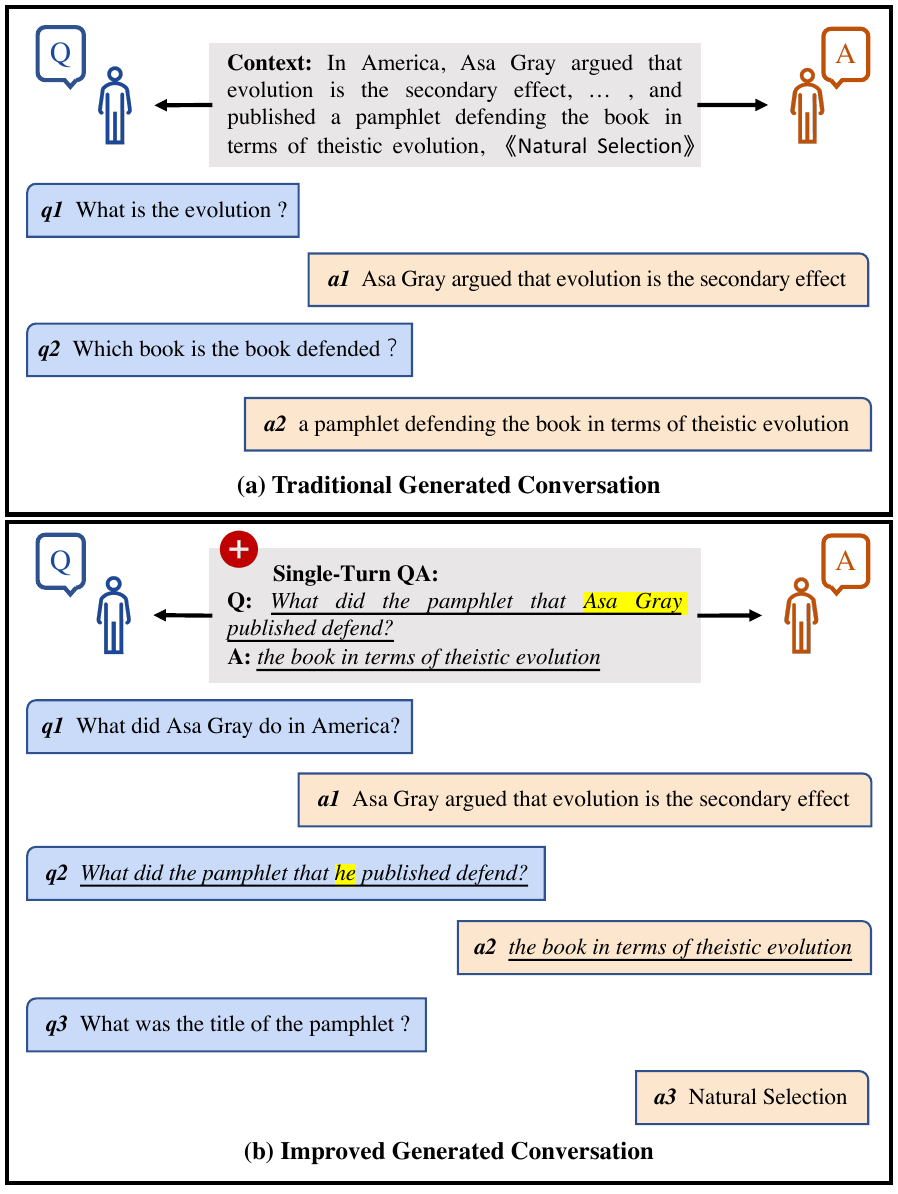}
    \caption{Two examples of generated conversations. The former represents traditional context-based conversation generation; the latter additionally considers the single-turn QA pair and rewrites it. Our method considers both context and additional single-turn question-answer pairs, thus generating more complete information.}
    \label{fig:demon_example}
\end{figure}

    Therefore, the existing literature mainly discusses a part of the whole process. On the one hand, research on conversational question generation is dedicated to generating follow-up questions~\cite{Pan_Li_Yao_Cai_Sun_2019,Qi_Zhang_Manning_2020}; on the other hand, research on conversational question answering aims to improve the accuracy of answers~\cite{gu-etal-2021-chaincqg,Qu_Yang_Qiu_Zhang_Chen_Croft_Iyyer_2019,RoR_2021}. As far as we know, only SIMSEEK~\cite{simseek2022} combines the two methods. In addition to the above research directions, some researchers have studied the generation of single-turn question-answer pairs. Different from these studies, we propose a method to better use  single-turn datasets, which alleviates the distribution shift between single-turn and multi-turn datasets and generates more complete conversations than SIMSEEK, as shown in Figure~\ref{fig:demon_example}.

    In this paper, we propose a method to convert single-turn datasets to multi-turn datasets. It consists of three modules, 
    the candidate single-turn QA pair Generator, the QA pairs Reassembler, and the conversation question Rewriter. Firstly, the Generator generates a large number of single-turn candidate QA pairs. Then, the Reassembler forms sequential QA pairs from the generated and existing QA pairs. Finally, the Rewriter converts these self-contained questions to questions related to the specific conversation.


    A conversational dataset called S2M has been generated by the proposed method and evaluated by the CQA task. We conducted unsupervised and supervised experiments on the challenging CQA benchmark QuAC~\cite{Choi_QuAC}. In the unsupervised case, our experimental results demonstrate the effectiveness of the synthetic conversations from S2M that improve performance of baselines and reduce the performance gap between the unsupervised setting and the supervised setting. In the case of supervision, with the help of S2M, our model has achieved state-of-the-art performance on the validation set and ranked 1st on the leaderboard test set. To further verify the quality of S2M, we conducted a human evaluation to compare S2M with the original dataset QuAC and other generated datasets~\cite{simseek2022,RGX2022}. The results have shown that the quality of the conversations in S2M is higher than others in terms of answer accuracy, context relevance, and overall adequacy.


    Our contributions are summarized as follows : 
 \begin{itemize}
     \item We verify defects in directly using existing datasets for data augmentation. Our results also indicate the possibility of converting single-turn to multi-turn. \textbf{As far as we know, this work is the first attempt to successfully convert single-turn datasets to multi-turn datasets.}
     \item We propose a new method to build a new dataset called S2M. Extensive numerical results have been conducted on the QuAC benchmark. It is worth mentioning that the proposed method obtains state-of-the-art performance on the validation set and ranks 1st on the benchmark test set.
     \item A human evaluation of S2M has also been conducted to show that the conversations in S2M are more popular than other synthetic datasets for annotators.
 \end{itemize}

\vspace{-0.4cm}
\section{Approach}

In this section, we first formalize the CQA task and introduce two types of mainstream methods to generate datasets. Then, we introduce our method with three components. In our method, we first introduce how to generate high-quality single-turn question-answer pairs. Next, we introduce how to construct and use a knowledge graph to obtain sequential question-answer pairs. Finally, we introduce the question Rewriter to obtain question-answer pairs that depend on the conversation. 

\subsection{Background}
\subsubsection{Task Formulation}
Let $C=[s_1, s_2, ..., s_N]$ denote the context, where the sentence $s_i = \{w_1, w_2,..., w_M\}$ contains $M_i$ tokens. Given a question $q_t$ with $t$ being the turn of given question, CQA is asked to answer it correctly from the context $C$ based on its history $H_{<t}=[(q_1, a_1), ..., (q_{t-1}, a_{t-1})]$.

\begin{figure}
    \centering
    \includegraphics[width=0.4\textwidth]{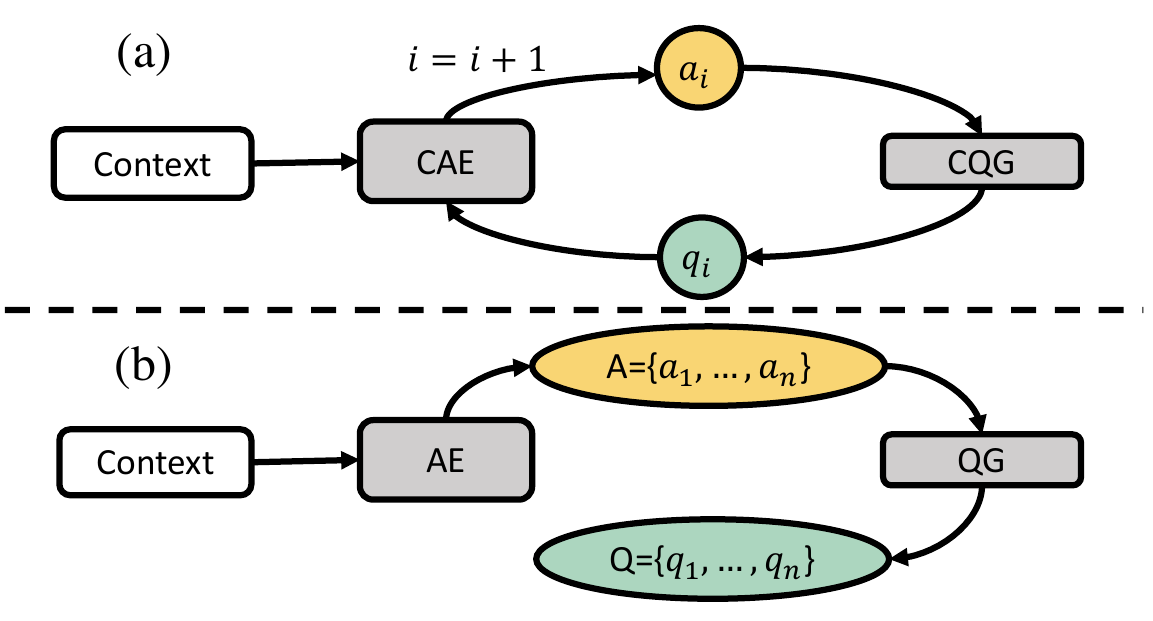}
    \caption{Two types of CQA dataset generation methods. Compared with method-b, method-a considers the conversational history. Although its generation is slow, the resulting quality is higher.}
    \label{fig:QG_framework}
\end{figure}

\vspace{-0.7cm}

\subsubsection{Data Generation Methods of CQA}

\begin{figure*}
    \centering
    \includegraphics[width=0.75\textwidth]{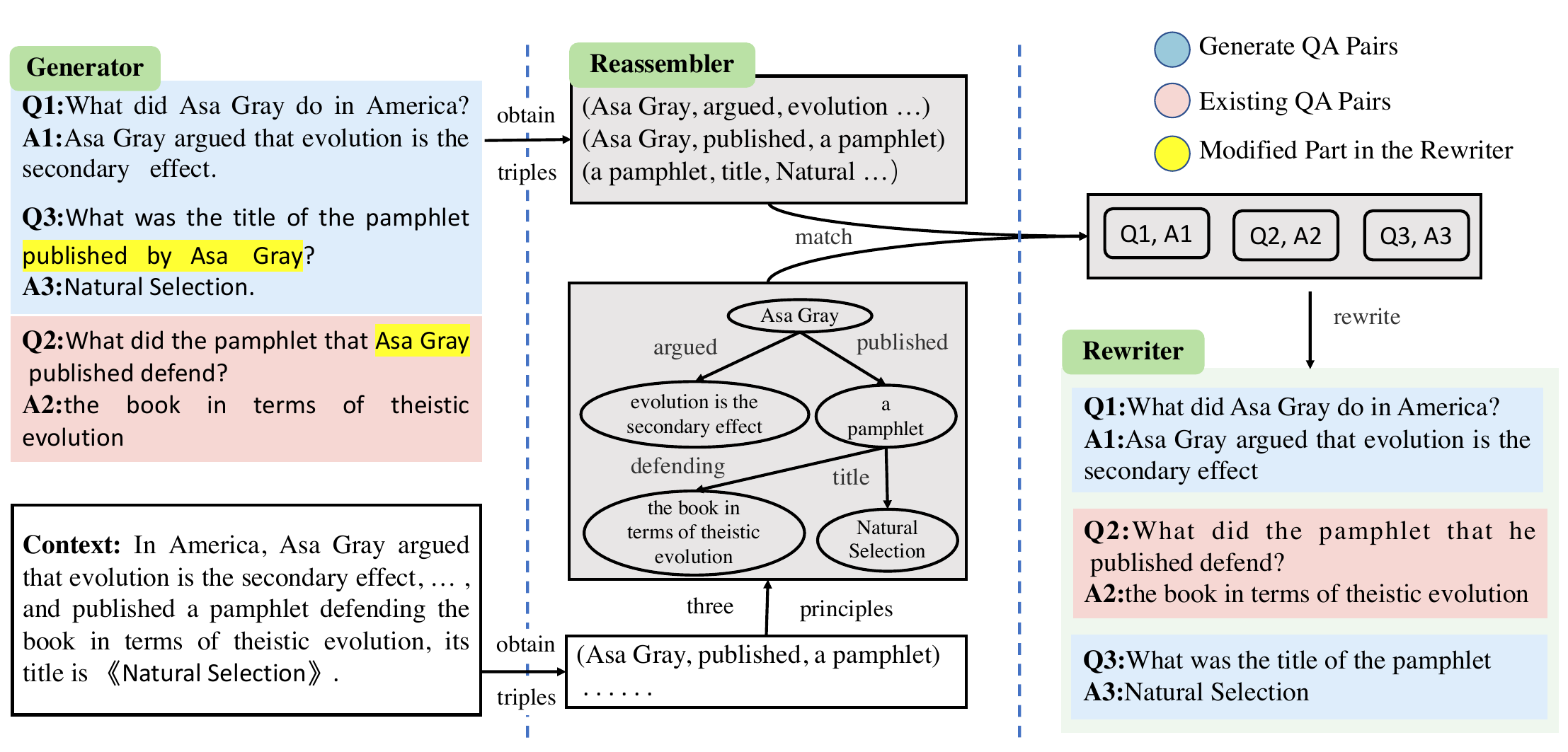}
    \caption{Our model overview consists of three main parts: a Single-Turn Candidate Question-Answer Pair, a QA pair Reassembler, and a Question Rewriter. The input to the model comprises the context and the existing single-turn question-answer pair of the red part. The output is the multi-turn dialogue rewritten by the Rewriter.}
    \label{fig:framework}
\end{figure*}


Although generating CQA datasets is a challenging task and there are a limited number of previously available works, the existing methods can be summarized into the following two types of methods, as shown in Figure~\ref{fig:QG_framework}.

In both methods, a context encoder encodes $C$ into a series of contextualized feature representations $\{h_i\}^L_{i=1}$, as shown below, where $L$ is the length of the input context.

\begin{equation}
    \mathbf{\{h_i\}_{i=1}^L} = ContextEncoder(C, H)
\end{equation}
$H$ represents the historical conversation information, which is optional and only effective in model-a of Figure~\ref{fig:QG_framework}. The context encoder $\textit{ContextEncoder}$ could be Long-Short Term Memory Network(LSTM) ~\cite{Hochreiter_Schmidhuber_2006} or pre-trained language models, e.g., BERT~\cite{Devlin_Chang_Lee_Toutanova_2019}. 

Second, to obtain answers $a_t$ for generating a question $q_t$, $\{h_i\}^L_{i=1}$ are projected onto start logit and end logit through multi-layer perceptrons separately. Both logits are then sent to a softmax function to compute the start and end probability distributions along all tokens in the context as shown in:

\begin{align}
    r^s_i &= \mathbf{W}^s_{2}\text{tanh}(\mathbf{}{W}^s_1\mathbf{h_i}) \\
    r^e_i &= \mathbf{W}^s_{2}\text{tanh}(\text{W}^e_1[\mathbf{h_i; h_s}]) \\
    p^s &= \text{softmax}(r^s) \\
    p^e &= \text{softmax}(r^e)
\end{align}
where $\mathbf{W}^s_1$, $\mathbf{W}^s_2$, $\mathbf{W}^e_1$, $\mathbf{W}^e_2$ are trainable parameters of the projection functions. $\mathbf{h}_s$ is the token representation of the start label, and ${p}^s \subseteq \mathbb{R}^L$, $p^{e} \subseteq \mathbb{R}^{L}$ are the start and the end probability distributions over all tokens, respectively. Based on the start and end logits, we obtain a set of candidate answers $\hat{A}_t = \{\hat{a}_t^1, \hat{a}_t^2, ... , \hat{a}_t^{k}\}$. We choose the result ${a}_t$ of the highest score as the answer to question $q_t$.

Third, after extracting the answer ${a}_t$, there are subtle differences between the two types of question-generation methods. On the one hand, the question ${q_t}$ is generated based on the extracted answer $a_t$ and context $C$, e.g. $p(q_t | C, {a}_t)$; on the other hand, the conversational history $H_{<t}$ is additionally considered when generating a question $q_t$, e.g. $p(q_t | C, {a}_t, H_{<t})$.

These methods have been widely proven to be effective in the CQA task. It is worth mentioning that both methods generate data based on contexts. If the context itself carries some single-turn question-answer pairs, they are powerless. Therefore, we propose the third type of method that converts single-turn datasets to multi-turn datasets. Next, we will introduce the three modules of the proposed method, as shown in Figure~\ref{fig:framework}.

\vspace{-0.2cm}
\subsection{Single-Turn Candidate Question-Answer Pair Generator}

Given a context $C$ in the single-turn dataset, we utilize the RGX ~\cite{RGX2022} framework to generate a large number of question-answer pairs and their corresponding credible scores $S_l$. The scores $S_l$ are denoted as the QAE loss:
\begin{align}
    S_l &= -(logP(I_{st}) + logP(I_{ed}))
\end{align}
where $P(I)$ is the probability of position $I$ in $C$. $I_{st}$ and $I_{ed}$ denote the start and end positions of every answer in the context $C$, respectively.

To get high-quality generated QA pairs, we evaluate them in terms of challenge and noise. Different from the RGX framework processing method, with the help of the EM algorithm~\cite{RGX2022}, we divide the QA pairs into four categories based on the credit scores of their questions: low, relatively low, medium, and high, respectively. The higher the level of QA pairs, the more challenging the questions in QA pairs and the more noisy the answers in QA pairs. The low indicates that the noise is low, but the challenge is also low. The high indicates that the challenge is high, but the noise is also high. We filter the low and high extreme QA pairs and select the remaining.

Since there is a lot of redundancy in the question-answer pairs generated by this framework, we propose the Improved Union Search algorithm, which merges redundant QA pairs and selects the QA pair with a medium score as their representative. We consider question-answer pairs with more than half of the total words repeated as redundant and add them to the redundant set. In the redundant set, the question-answer pair of the intermediate credit score is regarded as their representative.

\vspace{-0.3cm}

\subsection{Knowledge Graph for the Sequential Question-Answer Pairs}

In addition to the existing QA pairs in the single-turn dataset, we have obtained a large number of generated QA pairs. In this subsection, we will introduce how to construct a knowledge graph and how to use it to obtain sequential QA pairs. 

\vspace{-0.1cm}
\subsubsection{Constructing Knowledge Graph from the Context}
The knowledge graph is constructed based on structured information. To obtain the knowledge graph of a given context, we need to extract its triples. The current mainstream information extraction models~\cite{shang2022onerel,vasilkovsky2022detie,Yu_Sun_Cardie_Yu_2020} have shown that the open-source information extraction model is most suitable for the context from an arbitrary domain. Inspired by them, we choose the current state-of-the-art model OpenIE6 ~\cite{Kolluru_2020} to extract triples of the context.

\begin{figure}
    \centering
    \includegraphics[width=0.4\textwidth]{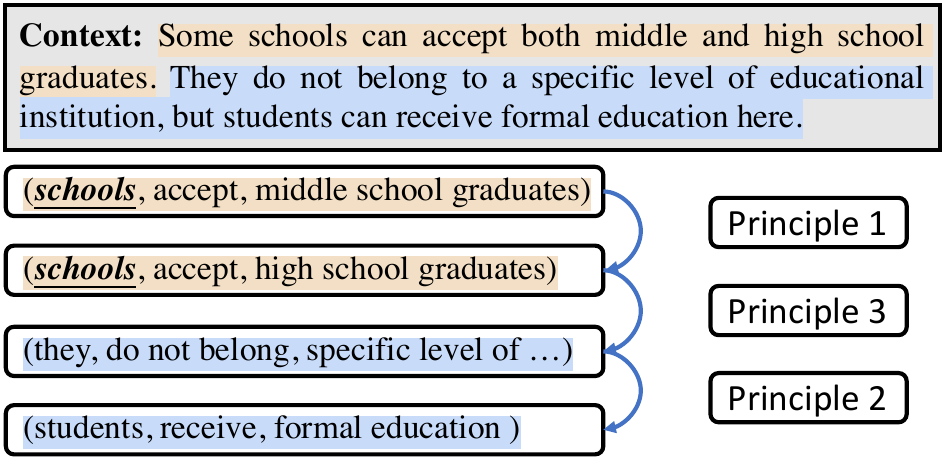}
    \caption{Example of triple concatenation with the three principles.}
    \label{fig:triple_sent}
\end{figure}

After extracting the triples of each sentence in the context, we realize that they are too complex to obtain the knowledge graph. As shown in Figure~\ref{fig:triple_sent}, the triples of adjacent sentences are not simply connected by the same head and tail. Therefore, we propose a \textbf{Triples Join Algorithm} based on the assumption that the triples of adjacent sentences are more semantically related. 
As shown in Figure~\ref{fig:triple_sent} and Algorithm~\ref{alg:triple join}, we connect triples based on the following three principles:
\vspace{-0.1cm}
 \begin{itemize}
     \item \textbf{Principle 1} The subject or object between two triples is the same or contained. For example, the subject or object of triple-A is equal to or contains the subject of triple-B.   
     \item \textbf{Principle 2} If there is an unconnected triple in the sentence, we connect it to an adjacent triple. 
     \item \textbf{Principle 3} If there is no connected triple between adjacent sentences, we connect the last triple of sentence $c_i$ with the first triple of sentence $c_{i+1}$.
 \end{itemize}

\vspace{-0.2cm}

\vspace{-0.4cm}

\begin{algorithm}[tb]
\caption{Triple Join Algorithm}
\label{alg:triple join}
\textbf{Input}: $C = [s_1, s_2, ... , s_n]$\\
\textbf{Parameter}: $s$ is a sentence of  context $C$. $n$ is the number of sentences in the context $C$\\
\textbf{Output}: Knowlege Graph $G$ 
\begin{algorithmic}[1] 
\STATE Initialize Graph $G$ = dict()
\STATE Initialize Triples $T$ = list()

\FOR{$s_i$ in $C_1^n$}
    \STATE $T$.append(OpenIE6($s_i$))
\ENDFOR

\STATE \textbf{\textcolor{red}{Intra-Sentence level}} \\
\textcolor{blue}{ /* The index k ranges from 1 to n. And the index i<>j  */ }
\FOR{$t_{k,i}, t_{k,j}$ in $T_k$}

    \IF{$t_{k,i}, t_{k,j}$ satisfy the Principle 1}
        \STATE connect($t_{k,i}, t_{k,j}$)
    \ENDIF
\ENDFOR

\FOR{$t_{k,i}, t_{k,i+1}$ in $T_{k}$}
    \IF{$t_{k,i}, t_{k,i+1}$ satisfy the Principle 2}
        \STATE connect($t_{k,i}, t_{k,i+1}$)
    \ENDIF
\ENDFOR

\STATE \textbf{\textcolor{red}{Inter-Sentence level}}
\FOR{$s_i, s_{i+1}$ in $C$}

    \IF{$s_i, s_{i+1}$ satisfy the Principle 3}
        \STATE connect($T_{i,end}, T_{i+1,start}$)
    \ENDIF
\ENDFOR
\STATE \textbf{connect(A,B)} \textcolor{blue}{connects the subjects and objects of the triples A \\and B. For example, $G[sub/obj\ of\ A].append(sub/obj\ of\ B)$.}

\STATE \textbf{return} $G$
\end{algorithmic}
\end{algorithm}


\vspace{-0.38cm}

\subsubsection{Obtaining Sequential QA Pairs from the Knowledge Graph}
After obtaining the knowledge graph representing the information flow of context, we use OpenIE6~\cite{Kolluru_2020} to extract triples from QA pairs as the primary information. The process of connecting sequential QA pairs in a coherent conversation can be thought of as performing a systematic walk over the Knowledge Graph~\cite{saha2018complex}. Based on the existing knowledge graph and triples, we obtain sequential QA pairs in two steps.

First, we match QA pairs and the knowledge graph. The nodes in the knowledge graph correspond to a triple. As shown by the Reassembler in Figure~\ref{fig:framework}, we mark the node when the knowledge graph node is the same as the triple in QA pairs. We match the generated and existing QA pairs with the knowledge graph, respectively. 


Second, we traverse the knowledge graph to obtain sequential QA pairs. We traverse all nodes from the root node of the knowledge graph and obtain all continuous masked nodes in order. We replace the mask nodes with matching QA pairs to obtain all sequential QA pairs of the context.

\vspace{-0.3cm}
\subsection{Question Rewriting for CQA}
Although the sequential QA pairs have a strong coherence between them, there is a lack of dialogue style between questions. Therefore, we convert self-contained questions to questions that depend on the conversation. We introduce how to rewrite the questions by the rewriting model in the following two steps.

\subsubsection{Obtaining the Questions Rewriting Dataset}
Specifically, we build our rewriting dataset R-CANARD based on CANARD~\cite{elgohary-etal-2019-unpack}. Given an instance ($C, H^c_{<t}, q^c_t, q_t$) in CANARD where $q^c_t$ and $q_t$ represent the follow-up question in one conversation and the self-contained question, respectively. $H^c_{<t}$ is the list of ($q^c_t, a_t$), representing the historical conversation of $q^c_t$, and $q_t$ is the rewriting target. According to the task, we construct an instance in R-CANARD as ($C, H_{<t}, q_t, q^c_t$), however $H_{<t}$ is the historical conversation of $q_t$, $q^c_t$ is our rewriting target. 

In traditional question rewriting methods, researchers modify follow-up questions so that they can be correctly interpreted outside of the conversational context~\cite{anantha_open2021}. Different from them, we reverse the question rewriting process by converting self-contained questions to questions that depend on the conversation. We build a new dataset and train a reverse question rewriting model for this task.

\subsubsection{Training the Conversational Question Rewriting Model}

Grounded on each self-contained question, the question rewriting model generates a follow-up question based on context. Thus, it should satisfy multiple objectives simultaneously, generating proper questions and aligning with historical conversations. Formally, the \textbf{C}onversational \textbf{Q}uestion \textbf{R}ewriting(CQR) model rewrites the self-contained question based on context and its history, e.g. $p^{CQR}_q(q^c_t | C, H_{<t}, q_t)$. 

\begin{figure}
    \centering
    \includegraphics[width=0.4\textwidth]{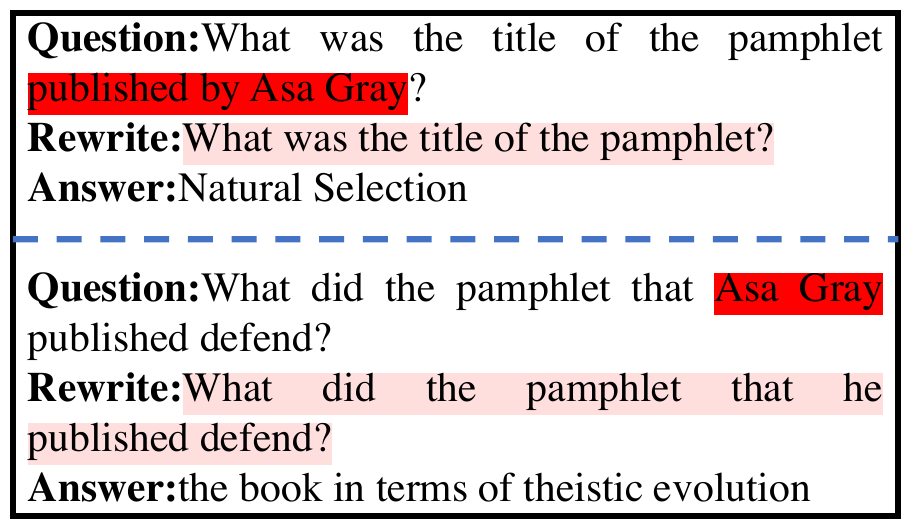}
    \caption{Two types of examples of rewritten questions.}
    \label{fig:convert_redundant}
\end{figure}

Now, we use the CQR model trained on the R-CANARD dataset to rewrite the questions in the sequential question-answer pairs. As shown in Figure~\ref{fig:convert_redundant}, we give two representative examples. First, we use pronouns to replace nouns that appear repeatedly between questions. Second, we omit the unimportant part of the question. Note that the rewritten questions are more realistic.


\vspace{-0.4cm}

\section{Experimental Setup}

In this section, we first introduce the datasets and baselines. Then, we define the evaluation metrics. Finally, we show the details of hyperparameters in our model.

\vspace{-0.2cm}
\subsection{Datasets}
To obtain synthetic conversations, we conduct experiments on one real-world dataset (i.e., SQuAD 2.0)~\cite{rajpurkar-etal-2018-know}. In the following, we use SQuAD to represent SQuAD2.0. Note that each instance in SQuAD  contains one context and countless single-turn QA pairs. To evaluate the effect of synthetic conversations, we train baselines on the recent CQA benchmark, QuAC~\cite{Choi_QuAC}, which consists of 100k QA multi-turn pairs. CANARD converts questions in QuAC to self-contained ones that can be understood without the conversation. Unlike CANARD, we construct the R-CANARD dataset, which converts self-contained questions in CANARD to the follow-up questions.

\subsection{Baselines for Synthetic CQA Generation}
We introduce solid baselines for synthesizing CQA datasets and compare them with our method. Due to the scarcity of previous work, we choose representative generation methods RGX and SIMSEEK as our baselines. 
 \begin{itemize}
     \item \textbf{RGX} This model is proposed by Luo~\cite{RGX2022}, one of the dominant frameworks in single-turn QA tasks. Compared with the traditional QA generation model, it leverages a self-training technique to improve the performance of both question generation and answer extraction.
     \item \textbf{SIMSEEK} The model proposed by Kim~\cite{simseek2022} is the previous state-of-the-art method for synthesizing CQA datasets. Compared with RGX, it considers the conversation history when generating QA pairs.
 \end{itemize}
 After building these synthetic conversations, we will use three different backbone architectures, ROBERTA-large, ElECTRA-large ,and DEBERTA-large, to verify the effectiveness of our method comprehensively.

\vspace{-0.1cm}

\subsection{Evaluation Metrics}
Two metrics are used to evaluate answer span prediction in the QuAC task: word-level F1 and human equivalence score (HEQ). The former measures the overlap between the predicted and actual spans while ignoring stopwords. The latter measures the percentage of examples for which the model's F1 score is higher than the average human F1 score. HEQ has two versions, HEQ-Q and HEQ-D. HEQ-Q calculates the percentage of questions for which the model's predictions exceed the human assessment, while HEQ-D calculates the percentage of conversations in which all questions exceed the human assessment.

\vspace{-0.1cm}
\subsection{Implementation Details}
We train Pre-trained Language Models (PLMs) in the following two stages. During training, we found that the order of training on the synthetic conversation and QuAC dataset affects the final results. The later the training order of QuAC is, the better the experimental results will be. Therefore, we first train PLMs on synthetic conversation, followed by the QuAC dataset. We report both the intermediate and final results, which correspond to the performance in the unsupervised and supervised environments, respectively.

\subsubsection{Hyperparameters} 
In the experiments, the max sequence length of questions is set to 128 and the answer length is set to 64. The stride of the sliding window for splitting documents is set to 128. The batch size is set to 16. The model is optimized using Adam, and we set the learning rates to 1e - 6, 1e-5, and 1e-5 for ROBERTA, ELECTRA, and DEBERTA, respectively. The random seed is always set to 42. In the inference process, we use beam search to predict the end position based on the start position, and the beam size is 5. All the other hyper-parameters are the same as reported in the corresponding papers. We run our experiments on $4\times$A100 GPUs. Training our models on synthetic conversations and the QuAC dataset takes about 12 hours.

To compare fairly with the previous best performance model SIMSEEK, we also employ another CQA dataset, CoQA, as an additional training resource. Without special instructions, we add the CoQA dataset when training by default.

\begin{table} \normalsize 
\centering 
\caption{Statistics of datasets. We record the number of conversations, questions, and their domains in the dataset.}
\label{table:datasets statics}
\begin{tabular}{lccc}
\hline
Dataset    & Domain               & Dialogs   & Ques.   \\ \hline
\multicolumn{4}{c}{\textbf{Existing single-turn QA}}                      \\
SQuAD      & Wikipedia            &           & 107K    \\ \hline
\multicolumn{4}{c}{\textbf{Existing multi-turn QA}}                                 \\
QuAC       & Wikipedia (People)   & 13K       & 98K     \\
CoQA       & 7 sub-domains        & 13K       & 127K    \\ \hline
\multicolumn{4}{c}{\textbf{Generated datasets}} \\
RGX        & Wikipedia            &           & 234K        \\
SIMSEEK    & Wikipedia            & 213K      & 2.1M    \\
S2M        & Wikipedia            & 19K          & 342K        \\ \hline
\end{tabular}
\end{table}

\vspace{-0.2cm}
\section{Main Results}

In this section, we introduce the construction process of synthetic conversations S2M and compare results with other datasets. Then, we conduct experiments on two datasets of the QuAC benchmark. First, we test the S2M dataset on three backbones and compare the results with other generated datasets on the QuAC validation set. Second, we test the S2M dataset on the QuAC benchmark test set.

\subsection{Dataset Construction}
Given the dataset SQuAD, we use our method to generate conversations based on the context and single-turn QA pairs. Specifically, we generate conversations until the twelfth turn or meet the termination condition. The termination conditions include the discontinuity of QA pairs and more than three unanswerable questions in one conversation. Table~\ref{table:datasets statics} shows the overall statistics of S2M and its comparison with other datasets. It can be seen that the dataset we generated is only one-seventh of the SIMSEEK dataset.

\begin{table} 
\caption{The results of the S2M dataset on different Pre-trained Language Models (PLMs) in QuAC. The experimental evaluation includes two settings: unsupervised, which means only fine-tuning in S2M, and supervised, which means additional fine-tuning in the QuAC training set.}
\label{table:plms result}
\resizebox{\linewidth}{!}{
\begin{tabular}{lcccccc}
\toprule[1.2pt]
\textbf{}       & \multicolumn{3}{c}{\textbf{Unsupervised}}   & \multicolumn{3}{c}{\textbf{Supervised}}       \\ \cmidrule(r){2-4} \cmidrule(r){5-7}
\textbf{Models} & \textbf{F1} & \textbf{HEQ-Q} & \textbf{HEQ-D} & \textbf{F1}   & \textbf{HEQ-Q} & \textbf{HEQ-D} \\ \toprule[1.2pt]
ROBERTA         & 12.6        & 2.6           & 0.1           & 73.4          & 69.5          & 13.1          \\
ROBERTA+S2M     & 48.0        & 41.2          & 1.5           & 74.5          & 71.6          & 15.4          \\ \hline \hline
ELECTRA         & 11.1        & 1.8           & 0.1           & 74.5          & 71.4          & 14.8          \\
ELECTRA+S2M     & 58.7        & 51.1          & 3.0           & 75.6          & 72.9          & 14.8          \\ \hline \hline
DEBERTA         & 13.2        & 2.5           & 0.0           & 74.7          & 72.0          & 14.7          \\
DEBERTA+S2M     & 59.2        & 51.5          & 3.0           & \textbf{76.4} & \textbf{73.5} & \textbf{16.2} \\ \bottomrule[1.2pt]
\end{tabular}
}

\end{table}

\vspace{-0.2cm}
\subsection{Results on QuAC}
We train PLMs on the S2M dataset and the QuAC dataset. Table~\ref{table:plms result} shows the results of the two-stage experiments on the QuAC development set. In the first stage, we trained on synthetic conversations, and the results showed that the performance of PLMs in the unsupervised setting was significantly improved. Specifically, through S2M, the DEBERTA's performance gap between the unsupervised setting and fully supervised setting is 15.5, which shows the authenticity of the synthetic conversations. In the second stage, we further fine-tuned models on the QuAC dataset, and the results showed that all PLMs increased by 1.3 on average, proving the generalizability of our method. Among all PLMs, DEBERTA performs the best, second only to ELECTRA.

Under the DEBERTA backbone, we further compare our generation method with other generation methods. Table~\ref{table:methods results} shows the performance of DEBERTA trained on the resulting datasets. While other methods can improve performance, there are some limitations compared to our method. DEBERTA-RGX shows the lowest performance. It implies the difficulty of directly extending generated single-turn QA methods to CQA. Adopting CQG modules that consider historical conversations (DEBERTA-SIMSEEK) improves CQA performance by more than 1.5 F1 scores. This indicates that understanding conversational questions is crucial for improving CQA performance. Although S2M is only one-seventh the size of the SIMSEEK dataset, DEBERTA-S2M achieves the highest score when considering both single-turn QA and conversational context. Specifically, DEBERTA-S2M outperforms the DEBERTA-SIMSEEK in terms of F1/HEQ-Q/HEQ-D, indicating that it is particularly effective for CQA despite the smaller dataset size.

\begin{table} \normalsize

\centering
\caption{Comparison of Automatic Generation Datasets on the QuAC Dataset.}
\label{table:methods results}
\begin{tabular}{lccc}
\toprule[1.2pt]
Models          & F1            & HEQ-Q         & HEQ-D         \\ \toprule[1.2pt]
DEBERTA         & 74.7          & 72            & 14.7          \\
DEBERTA-RGX     & 74.9($\uparrow0.2$)          & 72.1          & 14.1          \\
DEBERTA-SIMSEEK & 76.2($\uparrow1.5$) & 73.5          & 15.4          \\
DEBERTA-S2M     & \textbf{76.4($\uparrow1.7$)} & \textbf{73.5} & \textbf{16.2} \\ \bottomrule[1.2pt]
\end{tabular}

\end{table}

\vspace{-0.2cm}
\subsection{Official leaderboard results on QuAC}
QuAC challenge provides a hidden test set. Table~\ref{table:leaderboard result} displays the span prediction results of all baselines and our model. From this, we can see that our model DEBERTA-S2M outperforms the previous best performance model CDQ-DEBERTA and achieves new state-of-the-art performance on all three metrics. From the leaderboard results, we observe that the top-ranking models mainly use advanced pre-trained models and consider historical information. For example, CDQ-DEBERTA boosts the F1 score of ROR from 74.9 to 75.8 with the help of DEBERTA. Compared to BiDAF++, BiDAF++w/ 2-Context incorporates two turns of previous dialog history and significantly improves the performance of BiDAF++. 
We are the first to use generation datasets to help the CQA task and achieve success.
 
\begin{table} \normalsize

\centering
\caption{DEBERTA-S2M Model Ranking on the QuAC Benchmark Test Set.}
\label{table:leaderboard result}
\begin{tabular}{lccc}
\toprule[1.2pt]
Models               & \multicolumn{1}{l}{F1}   & \multicolumn{1}{l}{HEQ-Q} & \multicolumn{1}{l}{HEQ-D} \\
\toprule[1.2pt]
Human                & 81.1 & 100   & 100   \\ \hline \hline
DEBERTA-S2M          & \textbf{76.3}            & \textbf{73.6}             & \textbf{17.9}             \\ \hline \hline
CDQ-DEBERTA          & 75.8                     & 73.1                      & 15.9                      \\
AERmodel             & 75.2                     & 72.5                      & 16.5                      \\
RoR                  & 74.9                     & 72.2                      & 16.4                      \\
EL-QA                & 74.6                     & 71.6                      & 16.3                      \\
HistoryQA            & 74.2                     & 71.5                      & 13.9                      \\
TR-MT                & 74.4                     & 71.3                      & 13.6                      \\
HAM                  & 65.4                     & 61.8                      & 6.7                       \\
HAE                  & 62.4                     & 57.8                      & 5.1                       \\
BiDAF++ w/ 2-Context & 60.1                     & 54.8                      & 4.0                       \\
BiDAF++              & 50.2                     & 43.3                      & 2.2                       \\ \bottomrule[1.2pt]
\end{tabular}
\end{table}

\vspace{-0.2cm}

\section{Analysis}

\subsection{Qualitative Analysis}

\subsubsection{S2M is essential for data augmentation in the CQA task.} 

Table~\ref{table:result existing datasets} records the data augmentation results on the single-turn SQuAD, multi-turn CoQA, and S2M datasets. Here S2M does not contain the CoQA dataset, only for generating data. Although their contexts are all sourced from Wikipedia, the results are pretty different. In contrast, CoQA has a better boost than SQuAD. This is because both CoQA and QuAC are multi-turn conversation datasets. This situation illustrates the advantages of converting a single-turn dataset to a multi-turn dataset. On the other hand, CoQA is not as effective as S2M. This might be caused by the distribution shift between them and the target dataset, such as the shorter answer length of CoQA. Our method can be one of the solutions to mitigate the shift and provide further improvements. Furthermore, our method is a model-agnostic framework where all QG models can be adopted as a generator for obtaining conversation datasets.


\begin{table} \normalsize

\centering
\caption{Comparison of the Existing Datasets on the QuAC Dataset.}
\label{table:result existing datasets}
\begin{tabular}{lccc}
\toprule[1.2pt]
Models       & F1             & HEQ-Q & HEQ-D \\ \toprule[1.2pt]
DEBERTA-Large      & 74.7           & 72    & 14.7  \\
\quad + SQuAD      & 74.9($\uparrow0.2$)           & 72.1  & 14.1  \\
\quad + CoQA       & 75.1($\uparrow0.4$)           & 72.4  & 15.2  \\
\quad + S2M        & 75.6($\uparrow0.9$)          & 72.8 & 15.8  \\
\bottomrule[1.2pt]
\end{tabular}
\end{table}

\vspace{-0.5cm}

\subsubsection{Conversations in S2M are more relevant and coherent.} 

In Table~\ref{table:detail statics}, we compare the generated dialogues obtained by various methods with QuAC. These data show similarities in the token lengths of questions and answers. In contrast, S2M achieves higher scores at the word level f1, which measures how many words in the current and historical responses are reused in the current question, respectively. From the results, on the one hand, the question and answer in a QA pair from S2M are more correlated; on the other hand, the dialogue in S2M has better coherence.

\begin{table} \small
\centering
\caption{Statistical Analysis of Generated Datasets: A Comprehensive Study}
\label{table:detail statics}
\begin{tabular}{lrrr}
\toprule[1.5pt]
\multirow{2}{*}{} & \multirow{2}{*}{QuAC} & \multicolumn{2}{c}{Methods} \\
                  &                       & \scriptsize SIMSEEK         & \scriptsize S2M        \\ \hline
tokens / question & 6.5                   & 8.3            & 9.6        \\
tokens / answer   & 12.6                  & 14.6           & 15.3       \\ \hline
F1 of $(q_t, a_t)$      & 5.8                   & 7.3           & \textbf{33.4}       \\
F1 of $(q_t, a_{0:(t-1)})$      & 5.6                   & 8.6            & \textbf{10.8}       \\ \bottomrule[1.5pt]
\end{tabular}
\end{table}

\vspace{-0.5cm}

\subsection{Human Evaluation}
In this section, we conduct a human evaluation to detect the quality of synthetic conversations. Specifically, we employ in-house annotators to assess the quality of follow-up QA pairs. 

We hired five annotators to score the QA pairs of S2M and SIMSEEK, QuAC and S2M according to the four metrics inspired by SIMSEEK~\cite{simseek2022}. We let the annotators select 800 contexts from each dataset, obtaining about 5000 QA pairs. Samples were repeatedly scored by two annotators on the four metrics. The four metrics are detailed below:
\begin{itemize}
    \item \textbf{Overall adequacy} represents how adequate the QA pair is for continuing the conversation.
    \item \textbf{Informativeness} represents the amount of new information the question is trying to gather.
    \item \textbf{Context relevance} represents how relevant the question is to the given context.
    \item \textbf{Answer accuracy} represents whether the answer is accurate for the question.
\end{itemize}

\begin{figure}
    \centering
    \includegraphics[width=0.5\textwidth]{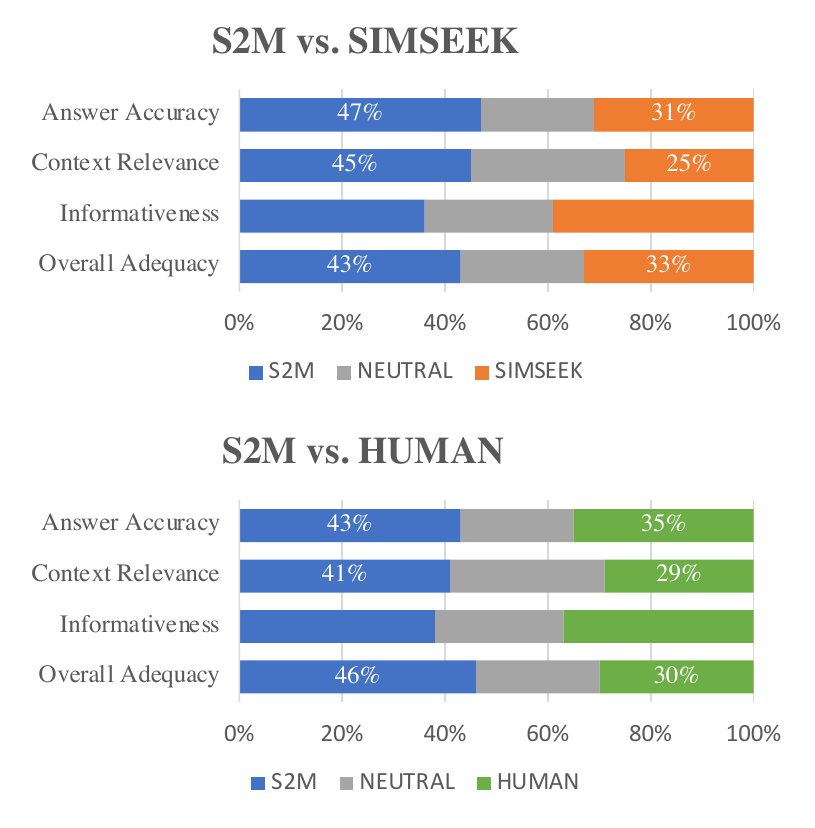}
    \caption{Human Evaluation of S2M, SIMSEEK, and QuAC Datasets: Comparing Overall Adequacy and Additional Metrics.}
    \label{fig:human evaluate}
\end{figure}

Figure~\ref{fig:human evaluate} compares S2M with other datasets using these metrics. We found that they are similar in informativeness. However, our method has better results in the remaining three metrics.

\vspace{-0.4cm}
\subsubsection{S2M benefits from the knowledge graph structure; they are more adequate.} 

Higher overall adequacy and context relevance have been observed in S2M. This is due to our context-based knowledge graph structure. It ensures that the conversation is advanced step by step according to context. In contrast, SIMSEEK relies entirely on pre-trained models and lacks guidance information, resulting in a break from context; QuAC requires concise responses, which seems relatively less enthusiastic. Therefore, our synthetic conversations are selected more frequently in terms of overall adequacy and context relevance.

\subsubsection{Conversations from S2M are reviewed as more accurate even than humans. } 

Furthermore, annotators conclude that S2M has higher accuracy. As shown in Figure~\ref{fig:human evaluate}, S2M is highly interactive between the answer and the current question, which makes it considered more helpful and communicative by annotators. 

In summary, our method successfully generates accurate QA pairs in each turn, which are more adequate. 

\vspace{-0.2cm}
\section{Related Work}

\textbf{Conversational Question Answering. }
With the release of large-scale CQA benchmarks~\cite{Choi_QuAC,reddy2019coqa}, more and more researchers are studying this challenging task. At the beginning, researchers made structural improvements~\cite{ijcai2020p171,huang2018flowqa,Qu_Yang_Qiu_Zhang_Chen_Croft_Iyyer_2019,Zhang_Yang_Zhao_2022,RoR_2021}, such as adding historical dialogues and cutting the context to obtain answers~\cite{Qu_Yang_Qiu_Zhang_Chen_Croft_Iyyer_2019}. With the popularity of large-scale pre-trained language models~\cite{Clark2020ELECTRA,he2023debertav,liu2019roberta}, more researchers used pre-trained language models and data augmentation methods, such as ROR~\cite{RoR_2021}. Later, due to the limited data scale, researchers automatically generate data to help the model learn external knowledge. Our method starts from the third paradigm and automatically generates data~\cite{simseek2022,RGX2022}, improving model performance and reducing labor costs.

\textbf{Data Augmentation.}
In the  CQA task, there are two types of data augmentation. The first method uses existing datasets~\cite{NQ_2019,Rajpurkar_SQuAD,reddy2019coqa} for data augmentation. This method is limited by data scale and distribution shift of the existing datasets~\cite{RoR_2021}. Directly using single-turn and multi-turn datasets will reduce task performance~\cite{RGX2022,simseek2022}; the second method generates datasets for the target task. However, the work is rare, and only SIMSEEK~\cite{simseek2022} has studied the field. They generate multi-turn conversation datasets. Unlike their methods, we convert a single-turn dataset to a multi-turn dataset. We consider both single-turn QA and dialogue contexts.

\textbf{Conversation Question Rewrite. }
Question rewriting(QR) technology has been successfully applied in various fields, such as question integration, question optimization, and question expansion~\cite{grbovic2015context,papakonstantinou1999query,rizvi2004extending}. More recently, QR has been shown to be helpful in the field of conversational question answering~\cite{anantha_open2021,hansen2020you}. QR was first introduced to the CQA task by Elgohary, and they released the CANARD dataset~\cite{elgohary-etal-2019-unpack}, which provides rewrites for the conversational questions from the QuAC dataset. Unlike CANARD, we try to apply QR to the conversational data generation task and propose the conversation question rewriting task, which rewrites the self-contained question as a question that depends on the conversation.

\section{Conclusion}
In this paper, we discussed the shortcomings of existing datasets for data augmentation in the field of conversation question answering. To solve these problems, we proposed a method of converting single-turn datasets to multi-turn datasets. Using this method, we constructed a high-quality S2M dataset and verified its performance on the validation set and test set of the public dataset QuAC. Notably, we ranked 1st in the QuAC benchmark. Finally, we compared our generation method with the previous state-of-the-art method from the qualitative analysis and human evaluation perspective. The results showed that our synthetic conversations have better results in terms of answer accuracy, context relevance, and overall adequacy. A potential research direction is to jointly optimize all components of the proposed method.

\section{Acknowledgements}
This work was supported in part by Ant Group. Any opinions, findings and conclusions or recommendations expressed in this material are those of the authors and do not necessarily reflect those of the sponsor.

\bibliography{ecai}

\begin{thebibliography}{10}

\bibitem{anantha_open2021}
Raviteja Anantha, Svitlana Vakulenko, Zhucheng Tu, Shayne Longpre, Stephen
  Pulman, and Srinivas Chappidi, `Open-domain question answering goes
  conversational via question rewriting', in {\em Proceedings of the 2021
  Conference of the North American Chapter of the Association for Computational
  Linguistics: Human Language Technologies}, pp. 520--534, Online, (June 2021).
  Association for Computational Linguistics.

\bibitem{ijcai2020p171}
Yu~Chen, Lingfei Wu, and Mohammed~J. Zaki, `Graphflow: Exploiting conversation
  flow with graph neural networks for conversational machine comprehension', in
  {\em Proceedings of the Twenty-Ninth International Joint Conference on
  Artificial Intelligence, {IJCAI-20}}, ed., Christian Bessiere, pp.
  1230--1236. International Joint Conferences on Artificial Intelligence
  Organization, (7 2020).
\newblock Main track.

\bibitem{Choi_QuAC}
Eunsol Choi, He~He, Mohit Iyyer, Mark Yatskar, Wen-tau Yih, Yejin Choi, Percy
  Liang, and Luke Zettlemoyer, `Quac: Question answering in context', in {\em
  Proceedings of the 2018 Conference on Empirical Methods in Natural Language
  Processing}, (Jun 2019).

\bibitem{Clark2020ELECTRA}
Kevin Clark, Minh-Thang Luong, Quoc~V. Le, and Christopher~D. Manning,
  `Electra: Pre-training text encoders as discriminators rather than
  generators', in {\em International Conference on Learning Representations},
  (2020).

\bibitem{Devlin_Chang_Lee_Toutanova_2019}
Jacob Devlin, Ming-Wei Chang, Kenton Lee, and Kristina Toutanova, `Bert:
  Pre-training of deep bidirectional transformers for language understanding',
  in {\em Proceedings of the 2019 Conference of the North}, (Jul 2019).

\bibitem{elgohary-etal-2019-unpack}
Ahmed Elgohary, Denis Peskov, and Jordan Boyd-Graber, `Can you unpack that?
  learning to rewrite questions-in-context', in {\em Proceedings of the 2019
  Conference on Empirical Methods in Natural Language Processing and the 9th
  International Joint Conference on Natural Language Processing
  (EMNLP-IJCNLP)}, pp. 5918--5924, Hong Kong, China, (November 2019).
  Association for Computational Linguistics.

\bibitem{grbovic2015context}
Mihajlo Grbovic, Nemanja Djuric, Vladan Radosavljevic, Fabrizio Silvestri, and
  Narayan Bhamidipati, `Context-and content-aware embeddings for query
  rewriting in sponsored search', in {\em Proceedings of the 38th international
  ACM SIGIR conference on research and development in information retrieval},
  pp. 383--392, (2015).

\bibitem{gu-etal-2021-chaincqg}
Jing Gu, Mostafa Mirshekari, Zhou Yu, and Aaron Sisto, `{C}hain{CQG}:
  Flow-aware conversational question generation', in {\em Proceedings of the
  16th Conference of the European Chapter of the Association for Computational
  Linguistics: Main Volume}, pp. 2061--2070, Online, (April 2021). Association
  for Computational Linguistics.

\bibitem{hansen2020you}
Victor Petr{\'e}n~Bach Hansen and Anders S{\o}gaard, `What do you mean
  ‘why?’: Resolving sluices in conversations', in {\em Proceedings of the
  AAAI Conference on Artificial Intelligence}, volume~34, pp. 7887--7894,
  (2020).

\bibitem{he2023debertav}
Pengcheng He, Jianfeng Gao, and Weizhu Chen, `De{BERT}av3: Improving de{BERT}a
  using {ELECTRA}-style pre-training with gradient-disentangled embedding
  sharing', in {\em The Eleventh International Conference on Learning
  Representations}, (2023).

\bibitem{Hochreiter_Schmidhuber_2006}
Sepp Hochreiter and Jürgen Schmidhuber, `Long short-term memory', {\em Neural
  Computation},  1735–1780, (May 2006).

\bibitem{huang2018flowqa}
Hsin-Yuan Huang, Eunsol Choi, and Wen-tau Yih, `Flowqa: Grasping flow in
  history for conversational machine comprehension', {\em In International
  Conference on Learning Representations}, (2018).

\bibitem{simseek2022}
Gangwoo Kim, Sungdong Kim, Kang~Min Yoo, and Jaewoo Kang, `Generating
  information-seeking conversations from unlabeled documents', {\em Proceedings
  of the 2022 Conference on Empirical Methods in Natural Language Processing},
  2362--2378, (2022).

\bibitem{Kolluru_2020}
Keshav Kolluru, Vaibhav Adlakha, Samarth Aggarwal, Mausam, and Soumen
  Chakrabarti, `Openie6: Iterative grid labeling and coordination analysis for
  open information extraction', in {\em Proceedings of the 2020 Conference on
  Empirical Methods in Natural Language Processing (EMNLP)}, (Nov 2020).

\bibitem{NQ_2019}
Tom Kwiatkowski, Jennimaria Palomaki, Olivia Redfield, Michael Collins, Ankur
  Parikh, Chris Alberti, Danielle Epstein, Illia Polosukhin, Jacob Devlin,
  Kenton Lee, Kristina Toutanova, Llion Jones, Matthew Kelcey, Ming-Wei Chang,
  Andrew~M. Dai, Jakob Uszkoreit, Quoc Le, and Slav Petrov, `Natural questions:
  A benchmark for question answering research', {\em Transactions of the
  Association for Computational Linguistics}, {\bf 7},  453–466, (Aug 2019).

\bibitem{PAQ_2021}
Patrick Lewis, Yuxiang Wu, Linqing Liu, Pasquale Minervini, Heinrich Küttler,
  Aleksandra Piktus, Pontus Stenetorp, and Sebastian Riedel, `Paq: 65 million
  probably-asked questions and what you can do with them', {\em Transactions of
  the Association for Computational Linguistics}, {\bf 9},  1098–1115, (Oct
  2021).

\bibitem{liu2019roberta}
Yinhan Liu, Myle Ott, Naman Goyal, Jingfei Du, Mandar Joshi, Danqi Chen, Omer
  Levy, Mike Lewis, Luke Zettlemoyer, and Veselin Stoyanov, `Roberta: A
  robustly optimized bert pretraining approach', {\em arXiv preprint
  arXiv:1907.11692}, (2019).

\bibitem{RGX2022}
Hongyin Luo, Shang-Wen Li, Mingye Gao, Seunghak Yu, and James Glass,
  `Cooperative self-training of machine reading comprehension', in {\em
  Proceedings of the 2022 Conference of the North American Chapter of the
  Association for Computational Linguistics: Human Language Technologies}, pp.
  244--257, Seattle, United States, (July 2022). Association for Computational
  Linguistics.

\bibitem{Pan_Li_Yao_Cai_Sun_2019}
Boyuan Pan, Hao Li, Ziyu Yao, Deng Cai, and Huan Sun.
\newblock Reinforced dynamic reasoning for conversational question generation,
  Jul 2019.

\bibitem{papakonstantinou1999query}
Yannis Papakonstantinou and Vasilis Vassalos, `Query rewriting for
  semistructured data', {\em ACM SIGMOD Record}, {\bf 28}(2),  455--466,
  (1999).

\bibitem{Qi_Zhang_Manning_2020}
Peng Qi, Yuhao Zhang, and Christopher~D. Manning, `Stay hungry, stay focused:
  Generating informative and specific questions in information-seeking
  conversations', in {\em Findings of the Association for Computational
  Linguistics: EMNLP 2020}, (Nov 2020).

\bibitem{Qu_Yang_Qiu_Zhang_Chen_Croft_Iyyer_2019}
Chen Qu, Liu Yang, Minghui Qiu, Yongfeng Zhang, Cen Chen, W.~Bruce Croft, and
  Mohit Iyyer, `Attentive history selection for conversational question
  answering', in {\em Proceedings of the 28th ACM International Conference on
  Information and Knowledge Management}, (Nov 2019).

\bibitem{rajpurkar2018}
Pranav Rajpurkar, Robin Jia, and Percy Liang, `Know what you don{'}t know:
  Unanswerable questions for {SQ}u{AD}', in {\em Proceedings of the 56th Annual
  Meeting of the Association for Computational Linguistics (Volume 2: Short
  Papers)}, pp. 784--789, Melbourne, Australia, (July 2018). Association for
  Computational Linguistics.

\bibitem{rajpurkar-etal-2018-know}
Pranav Rajpurkar, Robin Jia, and Percy Liang, `Know what you don{'}t know:
  Unanswerable questions for {SQ}u{AD}', in {\em Proceedings of the 56th Annual
  Meeting of the Association for Computational Linguistics (Volume 2: Short
  Papers)}, pp. 784--789, Melbourne, Australia, (July 2018). Association for
  Computational Linguistics.

\bibitem{Rajpurkar_SQuAD}
Pranav Rajpurkar, Jian Zhang, Konstantin Lopyrev, and Percy Liang, `Squad:
  100,000+ questions for machine comprehension of text', in {\em Proceedings of
  the 2016 Conference on Empirical Methods in Natural Language Processing},
  (Dec 2016).

\bibitem{reddy2019coqa}
Siva Reddy, Danqi Chen, and Christopher~D Manning, `Coqa: A conversational
  question answering challenge', {\em Transactions of the Association for
  Computational Linguistics}, {\bf 7},  249--266, (2019).

\bibitem{rizvi2004extending}
Shariq Rizvi, Alberto Mendelzon, Sundararajarao Sudarshan, and Prasan Roy,
  `Extending query rewriting techniques for fine-grained access control', in
  {\em Proceedings of the 2004 ACM SIGMOD international conference on
  Management of data}, pp. 551--562, (2004).

\bibitem{saha2018complex}
Amrita Saha, Vardaan Pahuja, Mitesh Khapra, Karthik Sankaranarayanan, and
  Sarath Chandar, `Complex sequential question answering: Towards learning to
  converse over linked question answer pairs with a knowledge graph', in {\em
  Proceedings of the AAAI conference on artificial intelligence}, volume~32,
  (2018).

\bibitem{shang2022onerel}
Yu-Ming Shang, Heyan Huang, and Xianling Mao, `Onerel: Joint entity and
  relation extraction with one module in one step', in {\em Proceedings of the
  AAAI Conference on Artificial Intelligence}, volume~36, pp. 11285--11293,
  (2022).

\bibitem{vasilkovsky2022detie}
Michael Vasilkovsky, Anton Alekseev, Valentin Malykh, Ilya Shenbin, Elena
  Tutubalina, Dmitriy Salikhov, Mikhail Stepnov, Andrey Chertok, and Sergey
  Nikolenko, `Detie: Multilingual open information extraction inspired by
  object detection', in {\em Proceedings of the AAAI Conference on Artificial
  Intelligence}, volume~36, pp. 11412--11420, (2022).

\bibitem{Yu_Sun_Cardie_Yu_2020}
Dian Yu, Kai Sun, Claire Cardie, and Dong Yu, `Dialogue-based relation
  extraction', in {\em Proceedings of the 58th Annual Meeting of the
  Association for Computational Linguistics}, (Jul 2020).

\bibitem{Zhang_Yang_Zhao_2022}
Zhuosheng Zhang, Junjie Yang, and Hai Zhao, `Retrospective reader for machine
  reading comprehension', {\em Proceedings of the AAAI Conference on Artificial
  Intelligence},  14506–14514, (Sep 2022).

\bibitem{RoR_2021}
Jing Zhao, Junwei Bao, Yifan Wang, Yongwei Zhou, Youzheng Wu, Xiaodong He, and
  Bowen Zhou.
\newblock Ror: Read-over-read for long document machine reading comprehension.,
  Nov 2021.

\end{thebibliography}
\end{document}